\documentclass[authoryear,12pt]{article}

\usepackage{url}
\usepackage{color}
\usepackage{multirow}
\usepackage{subfig}
\usepackage{graphicx}
\usepackage[english]{babel}
\usepackage{colortbl}
\usepackage{amsmath}
\usepackage{hyperref}
\usepackage{natbib}
\usepackage{textcomp}
\usepackage{wrapfig}
\usepackage{booktabs}
\usepackage{algorithm2e}
\usepackage{paralist}

\newcommand{\cbctt}{CB-CTT}
\newcommand{\pectt}{PE-CTT}

\DeclareMathSizes{10}{10}{10}{10}

\everymath=\expandafter{\the\everymath\displaystyle}

\newcommand{\ignore}[1]{}

\title{Feature-based tuning of simulated annealing applied to the curriculum-based course timetabling problem}

\author{Ruggero Bellio\\
\small{DIES, University of Udine, via Tomadini 30/A, I-33100, Udine, Italy}\\
\small{\texttt{ruggero.bellio@uniud.it}}\\ \\
Luca Di Gaspero, Sara Ceschia, Andrea Schaerf\\
\small{DIEGM, University of Udine, via delle Scienze 208, I-33100, Udine, Italy}\\
\small{\texttt{\{luca.digaspero|sara.ceschia|andrea.schaerf\}@uniud.it}}\\ \\
Tommaso Urli\\
\small{Optimisation Research Group, NICTA Canberra Research Lab}\\
\small{Tower A Level 3, 7 London Circuit, Canberra ACT 2601, Australia}\\
\small{\texttt{tommaso.urli@nicta.com.au}}\\ \\
}

\date{}

\begin{document}

\maketitle





  \begin{abstract}
    We consider the university course timetabling problem, which is one of the
    most studied problems in educational timetabling. In particular, we focus
    our attention on the formulation known as the \emph{curriculum-based course
    timetabling problem} (\cbctt{}), which has been tackled by many researchers
    and for which there are many available benchmarks.

    The contribution of this paper is twofold. First, we propose an effective
    and robust single-stage simulated annealing method for solving the problem.
    Secondly, we design and apply an extensive and statistically-principled
    methodology for the parameter tuning procedure.  The outcome of this
    analysis is a methodology for modeling the relationship between search
    method parameters and instance features that allows us to set the parameters
    for unseen instances on the basis of a simple inspection of the instance
    itself.
    Using this methodology, our algorithm, despite its apparent simplicity, has
    been able to achieve high quality results on a set of popular benchmarks.

    A final contribution of the paper is a novel set of real-world instances,
    which could be used as a benchmark for future comparison.
  \end{abstract}

\section{Introduction}

The issue of designing the timetable for the courses of the incoming term is a
typical problem that universities or other academic institutions face at each
semester. There are a large number of variants of this problem, depending on the
specific regulations at the involved institutions \citep[see,
e.g.,][]{King13,Lewi08,Scha99}.
Among the different variants, two in particular are now considered standard, and
featured at the international timetabling competitions ITC-2002 and ITC-2007
\citep{MSPM10}. These standard formulations are the \emph{Post-Enrollment Course
Timetabling} (\pectt{}) \citep{LePM07} and \emph{Curri\-cu\-lum-Based Course
Timetabling} (\cbctt{}) \citep{DiMS07}, which have received an appreciable
attention in the research community so that many recent articles deal with
either one of them.

The distinguishing difference between the two formulations is the origin of the
main constraint of the problem, i.e., the conflicts between courses that prevent
one from scheduling them simultaneously. Indeed, in \pectt{} the source of
conflicts is the actual student enrollment whereas in \cbctt{} the courses in
conflict are those that belong to the same predefined group of courses, or
curricula.  However, this is only one of the differences, which actually include
many other distinctive features and cost components. For example in \pectt{}
each course is a self-standing event, whereas in \cbctt{} a course consists of
multiple lectures.  Consequently the soft constraints are different: in \pectt{}
they are all related to events, penalizing late, consecutive, and isolated ones,
while in \cbctt{} they mainly involve curricula and courses, ensuring
compactness in a curriculum, trying to evenly spread the lectures of a course in
the weekdays, and possibly preserving the same room for a course.

In this work we focus on \cbctt{}, and we build upon our previous work on this
problem \citep{BeDS12}. The key ingredients of our method are
\begin{inparaenum}[\itshape i)\upshape]\item a fast single-stage Simulated
Annealing (SA) method, and \item a comprehensive statistical analysis
methodology featuring a principled parameter tuning phase\end{inparaenum}.
The aim of the analysis is to model the relationship between the most relevant
parameters of the solver and the features of the instance under consideration.
The proposed approach tackles the parameter selection as a classification
problem, and builds a rule for choosing the set of parameters most likely to
perform well for a given instance, on the basis of specific features.
The effectiveness of this methodology is confirmed by the experimental results
on two groups of validation instances. We are able to show that our method
compares favorably with all the state-of-the-art methods on the available
instances.

As an additional contribution of this paper, we extend the set of available
problem instances by collecting several new real-world instances that can be
added to the set of standard benchmarks and included in future comparison.

The source code of the solver developed for this work is publicly available at
\url{https://bitbucket.org/satt/public-cb-ctt}.

\section{Curriculum-based course timetabling}

The problem formulation we consider in this paper is essentially the version
proposed for the ITC-2007, which is by far the most popular. The detailed
formulation is presented in \citep{DiMS07}, however, in order to keep the paper
self-contained, we briefly report it also in the following.  Alternative
formulations of the problem, used in the experimental part of the paper
(Sect.~\ref{sec:other-formulations}), are described in \citep{BDDS12}.
Essentially, the problem consists of the following entities

\begin{description}
\item[\textbf{Days, Timeslots, and Periods.}] We are given a number of
  \emph{teaching days} in the week. Each day is split into a fixed number of
  \emph{timeslots}. A \emph{period} is a pair composed of a day and a timeslot.
\item[\textbf{Courses and Teachers.}] Each course consists of
  \textup{a fixed} number of \emph{lectures} to be scheduled in distinct
  periods, it is attended by a number of \emph{students}, and is taught by a
  \emph{teacher}. For each course, there is a minimum number of days across
  which the lectures of the course should be spread. Moreover, there are some
  periods in which the course cannot be scheduled (e.g., teacher's or students'
  availabilities).
\item[\textbf{Rooms.}] Each \emph{room} has a \emph{capacity}, i.e., a number of
  available seats.
\item[\textbf{Curricula.}] A \emph{curriculum} is a group of courses that share
  common students. Consequently, courses belonging to the same curriculum are
  \emph{in conflict} and cannot be scheduled at the same period.
\end{description}

A solution of the problem is an assignment of a period (day and timeslot) and a
room to all lectures of each course so as to satisfy a set of \emph{hard}
constraints and to minimize the violations of \emph{soft} constraints described
in the following.

\subsection{Hard constraints}

There are three types of hard constraints

\begin{description}
\item[\textsf{RoomOccupancy}:] Two lectures cannot take place simultaneously in
  the same room.
\item[\textsf{Conflicts}:] Lectures of courses in the \emph{same curriculum},
  or \emph{taught by the same teacher}, must be scheduled in different periods.
\item[\textsf{Availabilities}:] A course may not be available for being scheduled
  in a certain period.
\end{description}

\subsection{Soft constraints}

For the formulation ITC-2007 there are four types of soft constraints

\begin{description}
\item[\textsf{RoomCapacity}:] The capacity of the
  room assigned to each lecture must be greater than or equal to the number of
  students attending the corresponding course. The penalty for the violation of
  this constraint is measured by the number of students in excess.

\item[\textsf{MinWorkingDays}:] The lectures of each course cannot be
  tightly packed, but they must be spread into a given minimum number of days.

\item[\textsf{IsolatedLectures}:] Lectures belonging to a curriculum
  should be adjacent to each other (i.e., be assigned to consecutive periods).
  We account for a violation of this constraint every time, for a given
  curriculum, there is one lecture not adjacent to any other lecture of the same
  curriculum within the same day.

\item[\textsf{RoomStability}:] All lectures of a course should be given
  in the same room.
\end{description}

For all the details, including  input and output data formats and validation
tools, we refer to \citep{DiMS07}.

\section{Related work}

In this section we review the literature on \cbctt{}.  The presentation is
organized as follows: we firstly describe the solution approaches based on
metaheuristic techniques; secondly, we report the contributions on exact
approaches and on methods for obtaining lower bounds; finally, we discuss papers
that investigate additional aspects related to the problem, such as instance
generation and multi-objective formulations.  A recent survey covering all these
topics is provided by \cite{BCRT15}.

\subsection{Metaheuristic approaches}

\cite{Mull09} solves the problem by applying a constraint-based solver that
incorporates several local search algorithms operating in three stages: (i) a
construction phase that uses an \emph{Iterative Forward Search} algorithm to
find a feasible solution, (ii) a first search phase delegated to a \emph{Hill
Climbing} algorithm, followed by (iii) a \emph{Great Deluge} or \emph{Simulated
Annealing} strategy to escape from local minima.
The algorithm was not specifically designed for \cbctt{} but it was intended to
be employed on all three tracks of ITC-2007 (including, besides \cbctt{} and
\pectt{}, also Examination Timetabling). The solver was the winner of two out of
three competition tracks, and it was among the finalists in the third one.

The \emph{Adaptive Tabu Search} proposed by \cite{LuHa09} follows a three stage
scheme: in the initialization phase a feasible timetable is built using a fast
heuristic; then the intensification and diversification phases are alternated
through an adaptive tabu search in order to reduce the violations of soft
constraints.

A novel hybrid metaheuristic technique, obtained by combining
\emph{Electro\-magne\-tic-like Mechanisms} and the \emph{Great Deluge}
algorithm, was employed by \cite{ATMM10} who obtained high-quality results on
both \cbctt{} and \pectt{} testbeds.

Finally, \cite{LuJG11} investigated the search performance of different
neighborhood relations typically used by local search algorithms to solve this
problem. The neighborhoods are compared using different evaluation criteria, and
new combinations of neighborhoods are explored and analyzed.

\subsection{Exact approaches and methods for computing lower bounds}

{
\renewcommand{\arraystretch}{1.1}
\renewcommand{\tabcolsep}{1pt}

\begin{table*}[htdp]
\caption{Lower bounds for the \texttt{comp} instances. Proven optimal solutions
in the \emph{Best known} column are denoted by an asterisk. \textbf{Note:} These
bounds were obtained using diverse experimental setups (particularly time budgets)
and therefore should not be regarded as a comparison of methods to find lower
bounds but rather as a record of the lower bounds available in literature.}
\begin{footnotesize}
\begin{center}
\begin{tabular}{lrrrrrrr>{\hspace{6ex}}r@{}l}
\toprule
&
\multicolumn{1}{c}{L\"u}&
\multicolumn{1}{c}{Burke} &
\multicolumn{1}{c}{Hao} &
\multicolumn{1}{c}{Lach} &
\multicolumn{1}{c}{Burke} &
\multicolumn{1}{c}{As\'{\i}n Ach\'a} &
\multicolumn{1}{c}{Cacchiani}  &
\multicolumn{2}{c}{Best}
\\
Instance &
\multicolumn{1}{c}{~~(2009)~~}&
\multicolumn{1}{c}{~~(2010)~~} &
\multicolumn{1}{c}{~~(2011)~~} &
\multicolumn{1}{c}{~~(2012)~~} &
\multicolumn{1}{c}{~~(2012)~~} &
\multicolumn{1}{c}{~~(2012)~~} &
\multicolumn{1}{c}{~~(2013)~~}  &
\multicolumn{2}{c}{known}
\\
\midrule
\texttt{comp01} & 4             & \textbf{5} & 4               & 4             & 4     & 0             & \textbf{5}    & 5 & *     \\ 
\texttt{comp02} & 11            & 6          & 12              & 11            & 0     & \textbf{16}   & \textbf{16}   & 24            \\ 
\texttt{comp03} & 25            & 43         & 38              & 25            & 2     & 28            & \textbf{52}   & 66            \\ 
\texttt{comp04} & 28            & 2          & \textbf{35}     & 28            & 0     & \textbf{35}   & \textbf{35}   & 35 & * \\ 
\texttt{comp05} & 108           & 183        & 183             & 108           & 99    & 48            & 166           & 284           \\ 
\texttt{comp06} & 12            & 6          & 22              & 10            & 0     & \textbf{27}   & 11            & 27  & *           \\ 
\texttt{comp07} & \textbf{6}    & 0          & \textbf{6}      & \textbf{6}    & 0     & \textbf{6}    & \textbf{6}    & 6 & *    \\ 
\texttt{comp08} & \textbf{37}   & 2          & \textbf{37}     & \textbf{37}   & 0     & \textbf{37}   & \textbf{37}   & 37 & *   \\ 
\texttt{comp09} & 47            & 0          & 72              & 46            & 0     & 35            & 92            & 96            \\ 
\texttt{comp10} & \textbf{4}    & 0          & \textbf{4}      & \textbf{4}    & 0     & \textbf{4}    & 2             & 4 & *    \\ 
\texttt{comp11} & 0             & 0          & 0               & 0             & 0     & 0             & 0             & 0 & *    \\ 
\texttt{comp12} & 57            & 5          & \textbf{109}    & 53            & 0     & 99            & 100           & 298           \\ 
\texttt{comp13} & 41            & 0          & \textbf{59}     & 41            & 3     & \textbf{59}   & 57            & 59 & *   \\ 
\texttt{comp14} & 46            & 0          & \textbf{51}     &               & 0     & \textbf{51}   & 48            & 51 & *  \\ 
\texttt{comp15} &               &            & \textbf{38}     &               &       & 28            & 52            & 66            \\ 
\texttt{comp16} &               &            & 16              &               &       & \textbf{18}   & 13            & 18 & *  \\ 
\texttt{comp17} &               &            & 48              &               &       & \textbf{56}   & 48            & 56 & *  \\ 
\texttt{comp18} &               &            & 24              &               &       & \textbf{27}   & 52            & 62            \\ 
\texttt{comp19} &               &            & \textbf{56}     &               &       & 46            & 48            & 57            \\ 
\texttt{comp20} &               &            & 2               &               &       & \textbf{4}    & \textbf{4}    & 4 & *    \\ 
\texttt{comp21} &               &            & \textbf{61}     &               &       & 42            & 42            & 74            \\ 
\bottomrule
\end{tabular}
\end{center}
\end{footnotesize}
\label{tab:LB}
\end{table*}%
}

Several authors tackled the problem by means of exact approaches with both the
goal of finding solutions or computing lower bounds.

Among these authors, \citet{BMPR10b} engineered a hybrid method based on the
decomposition of the whole problem into different sub-problems, each of them
solved using a mix of different IP formulations. Subsequently, the authors
propose a novel IP model \citep{BMPR10} based on the concept of ``supernode''
which was originally employed to model graph coloring problems. This new
encoding has been applied to the \cbctt{} benchmarks and was compared with the
established two-index IP model. The results showed that the supernodes
formulation is able to considerably reduce the computational time.  Lastly, the
same authors propose a \emph{Branch-and-Cut} procedure \citep{BMPR12} aimed at
computing lower bounds for various problem formulations.

\citet{LaLu12} proposed an IP approach that decomposes the problem in two
stages: the first aims at assigning courses to periods and mainly focuses on
satisfying the hard constraints; the second optimizes for the soft constraints
and assigns lectures to rooms by solving a matching problem.

\citet{HaBe11} developed a partition-based approach to compute lower bounds: The
idea behind their method is to divide the original instance into sub-instances
by means of an \emph{Iterative Tabu Search} procedure. Afterwards, each
subproblem is solved via an IP solver using the model proposed by
\citet{LaLu12}; a lower bound of the original instance is obtained by summing up
the lower bound values of all these sub-instances.

A somewhat similar approach has been tried recently by \citet{CCRT13}, who
instead exploit soft constraints for the sub-instance partitioning. After the
partitioning phase, two separated problems are formulated as IPs and then solved
to optimality by a \emph{Column Generation} technique.

\citet{AsNi14} employed several satisfiability (SAT) models for tackling the
\cbctt{} problem that differ on which constraints are specifically considered
soft or hard. Using different encodings they were able to compute lower bounds
and obtain new best solutions for the benchmarks.

Finally, \citet{MSTIS13} translated the \cbctt{} formulation into an Answer Set
Programming (ASP) problem and solved it using the ASP solver \texttt{clasp}.

A summary of the results of the aforementioned literature is given in
Table~\ref{tab:LB}, which reports the lower bounds for the instances of ITC-2007
testbed (named \texttt{comp} in the table). The results highlighted in boldface
indicate the tightest lower bounds.

Note that these bounds were obtained using diverse experimental setups
(particularly time budgets) and therefore should not be regarded as a fair
comparison of methods, but rather as a record of the lower bounds available in
literature. In the table we also report the best results known at the time of
writing. Best values marked with an asterisk are guaranteed optimal (i.e., they
match the lower bound).

\subsection{Additional research issues}

We now discuss research activities on \cbctt{} that consider other issues
besides the solution to the original problem.

The first issue concerns the development of instance generators. To the best of
our knowledge, the first attempt to devise an instance generator for \cbctt{} is
due to \citet{BMPR10}, who tried to create instances that resemble the structure
of the ITC-2007 \texttt{comp} instances. That generator has been revised and
improved by \citet{LoMi10}, who based their work on a deeper insight on the
instance features and made their generator publicly available. For our
experimental analysis we use the generator developed in the latter work.

A further research issue concerns the investigation of the problem as a
multi-objective one.  To this regard, \citet{Geig09} considered the \cbctt{}
problem as a multi-criteria decision problem, and studied the influence of the
weighted-sum aggregation methodology on the performance of the technique.

Another issue related to multi-objective studies concerns the concept of
\emph{fairness}. Notably, in the standard single objective formulation, some
curricula can be heavily penalized in the best result in favor of an overall
high quality.  \citet{MuWa13} studied this problem and considered various
notions of fairness. Moreover, they compared these notions in order to evaluate
their effects with respect to the other objectives.

\section{Search method}
\label{sec:search}

We propose a solution method for the problem that is based on the well-known
\emph{Simulated Annealing} metaheuristic \citep{KiGV83}. As it is customary for
local search, in order to instantiate the abstract solution method to the
problem at hand we must define a search space, a neighborhood relation, and a
cost function.

\subsection{Search space} We consider the search space composed of all the
assignments of lectures to rooms and periods for which the hard constraint
\textsf{Availability} is satisfied.  Hard constraints \textsf{Conflicts} and
\textsf{RoomOccupancy} are considered as components of the cost function and
their violation is highly penalized.

\subsection{Neighborhood relations} We employ two different basic neighborhood
relations, defined by the set of solutions that can be reached by applying any
of the following moves to a solution

\begin{description}
\item[\textsf{MoveLecture (ML)}:] Change the period and the room of one lecture.
\item[\textsf{SwapLectures (SL)}:] Swap the period and the room of two lectures
  of distinct courses.
\end{description}

The two basic neighborhoods are merged in a composite neighborhood that
considers the set union of \textsf{ML} and \textsf{SL}. However, according to
the results of our previous study \citep{BeDS12}, we restrict the \textsf{ML}
neighborhood so that only moves which place the lecture in an \emph{empty} room
are considered (whenever possible, i.e., when the room occupancy factor is less
than $100\%$, as there would be no empty rooms in such a case).
Moreover, our previous findings suggest employing a non-uniform probability for
selecting the moves in the composed neighborhood.  In detail, the move selection
strategy consists of two stages: first the neighborhood is randomly selected
with a non-uniform probability controlled by a parameter $sr$ (\emph{swap
rate}), then a random move in the selected neighborhood is uniformly drawn.

\subsection{Cost function} \label{sec:cost}
The cost of a solution $s$ is a weighted sum of the violations of the hard
constraints \textsf{Conflicts} and \textsf{RoomOccupancy} and of the objectives
(i.e., the measure of soft constraints violations)

\begin{align}
    \label{eqn:cost}
    Cost(s) & = \mathsf{Conflicts}(s) & \times\quad & w_{hard} \nonumber \\
            & + \mathsf{RoomOccupancy}(s) & \times\quad &  w_{hard} \nonumber \\
            & + \mathsf{MinWorkingDays}(s) & \times\quad &  w_{wd} \nonumber \\
            & + \mathsf{RoomStability}(s) & \times\quad & w_{rs} \nonumber \\
            & + \mathsf{RoomCapacity}(s) & \times\quad & w_{rc} \nonumber \\
            & + \mathsf{IsolatedLectures}(s) & \times\quad &  w_{il} \nonumber \, ,
\end{align}
where the weights $w_{wd} = 5$, $w_{rs} = w_{rc} = 1$ and $w_{il} = 2$ are
defined by the problem formulation, while $w_{hard}$ is a parameter of the
algorithm (see Section~\ref{sec:parameters}). This value should be high enough
to give precedence to feasibility over objectives, but it should not be so high
as to allow the SA metaheuristic (whose move acceptance criterion is based on
the difference in the cost function between the current solution and the
neighboring one) to select also moves that could increase the number of
violations in early stages of the search.

\subsection{The SA metaheuristic} \label{sec:metaheuristic}

In a departure from our previous work \citep{BeDS12}, in which the metaheuristic
that guides the search is a combination (token-ring) of Tabu Search and a
``standard'' version of SA, in this work we employ a single-stage enhanced
version of the latter method.  We show that this rather simple SA variant, once
properly tuned, outperforms such a combination.

The SA metaheuristic \citep{KiGV83} is a local search method based on a
probabilistic acceptance criterion for non-improving moves. Specifically, at
each iteration a neighbor of the current solution is randomly selected and it is
accepted if either\begin{inparaenum}[\itshape i)\upshape]\item it improves the
cost function value or, \item according to an exponential probability
distribution that is biased by the amount of worsening and by a parameter $T$
called temperature\end{inparaenum}.
Besides the initialization functions (setting the initial solution and fixing
temperature $T_{0}$), the main hot-spots of the method are the function for
updating (i.e., decreasing) the temperature and the stopping condition. In the
standard variant of SA, the temperature update is performed at regular intervals
(i.e., every $n_{s}$ iterations) and the cooling scheme employed is a geometric
one. That is, the temperature is decreased by multiplying it for a cooling
factor $\alpha$ ($0 < \alpha < 1$), so that $T' = \alpha \cdot T$. The search is
stopped when the temperature reaches a minimum value $T_{min}$ that will prevent
accepting worsening solutions.
%
%
The main differences between the SA approach implemented in this work and the
one proposed in \citep{BeDS12} reside in a different specification  of these two
components. In this implementation we replace the standard functions with the
following strategies

\begin{enumerate}
\item \label{search:cutoff} a \emph{cutoff-based} temperature cooling
  scheme \citep{JAMS89};
\item \label{search:stopping} a different \emph{stopping condition}
  for the solver, based on the maximum number of allowed iterations.
\end{enumerate}

In the following, we detail these two aspects of the SA method employed in this
work.

\paragraph{Cutoff-based cooling scheme}
In order to better exploit the time at its disposal, our algorithm employs a
cutoff-based cooling scheme. In practice, instead of sampling a fixed number
$n_s$ of solutions at each temperature level (as it is customary in SA
implementations), the algorithm is allowed to decrease the temperature
prematurely by multiplying it by a cooling rate $cr$, provided that a portion
$n_a \leq n_s$ of the \emph{sampled} solutions has been \emph{accepted} already.
This allows us to speed-up the initial part of the search, thus saving
iterations that can be used in the final part, where intensification takes
place.

\paragraph{Stopping condition}

To allow for a fair comparison with the existing literature, instead of
stopping the search when a specific (minimum) temperature $T_{min}$ is reached,
our algorithm stops on the basis of an iteration budget, which is roughly
equivalent to fixing a time budget, given that the cost of one iteration is
approximately constant.

A possible limitation of this choice is that the temperature might still be too
high when the budget runs out. In order to overcome this problem, we fix an
\emph{expected} minimum temperature $T_{min}$ to a reasonable value and we
compute the number $n_s$ (see Equation~\ref{eqn:neighbors_sampled}) of solutions
sampled at each temperature so that the minimum temperature is reached exactly
when the maximum number of iterations is met. That is

\begin{equation}
    \label{eqn:neighbors_sampled}
    \qquad n_s = iter_{max} \Bigg/ \left(\frac{-\log\left(T_{0}/T_{min}\right)}{\log cr}\right) .
\end{equation}

Because of the cutoff-based cooling scheme, at the beginning of the search the
temperature might decrease before all $n_s$ solutions have been sampled. Thus
$T_{min}$ is reached in $k$ iterations in advance, where $k$ depends on the cost
landscape and on the ratio $n_a/n_s$. These spared iterations are exploited at
the end of the search, i.e., after $T_{min}$ has been reached, to carry out
further intensification moves.

In order to simplify the parameters of the algorithm, given the dependence of
the cutoff-based scheme on the ratio $n_a / n_s$ and the relation $n_a \leq
n_s$, we decided to indirectly specify the value of the parameter $n_a$ by
including a real-valued parameter $\rho \in [0, 1]$ defined as $\rho = n_a /
n_s$.

\subsection{SA parameters}
\label{sec:parameters}

Our SA metaheuristic involves many parameters. In order to refrain from making
any ``premature commitment'' \citep[see][]{Hoos12}, we consider all of them in
the experimental analysis. They are summarized in Table~\ref{tab:parameters},
along with the ranges involved in the experimental analysis which have been
fixed based on preliminary experiments.

The iterations budget has been fixed, for each instance, to $iter_{max} = 3
\cdot 10^8$, a value that provides the algorithm with a running time which is
approximately equivalent to the one allowed by ITC-2007 computation rules
(namely 408 seconds on our test machine).

{
    \renewcommand{\arraystretch}{1.1}
    \renewcommand{\tabcolsep}{4pt}

    \begin{table}
        \caption{Parameters of the search method\label{tab:parameters}.}
        \centering
        \begin{tabular}{llc}
            \toprule
            Parameter                 & Symbol & Range \\
            \midrule
            Starting temperature               & $T_{0}$        & $[1, 100]$ \\
            Neighbors accepted ratio ($n_a/n_s$) & $\rho$        & $[0.01, 1]$ \\
            Cooling rate                       & $cr$           & $[0.99, 0.999]$     \\
            Hard constraints weight            & $w_{hard}$     & $[10, 1000]$     \\
            Neighborhood swap rate             & $sr$           & $[0.1, 0.9]$          \\
            Expected minimum temperature       & $T_{min}$      & $[0.01, 1]$           \\
            \bottomrule
        \end{tabular}
    \end{table}
}

\section{Problem instances}
\label{sec:instances}

We now describe the instances considered in this work and how we use them in our
experimental analysis. In particular, according to the customary
\emph{cross-validation} guidelines \citep[e.g.,][]{HaTF2009}, we have split the
instance set in three groups: a set of \emph{training instances} used to tune
the algorithm, a set of \emph{validation instances} used to evaluate the tuning,
and finally a set of \emph{novel instances} to verify the quality of the
proposed method on previously unseen instances.

\subsection{Training instances}
\label{sec:artificial-instances}

The first group of instances is a large set of artificial instances created
using the generator by Leo Lopes \cite[see][]{LoMi10}, which has been
specifically designed to reproduce the features of real-world instances.


The generator is parametrized on two instance features: the total number of
lectures and the percentage of room occupation.

In order to avoid \emph{overtuning}, this is the only group of instances that
has been used for the tuning phase. Accordingly, reporting individual results on
these instances it would be of little significance and are therefore omitted.
Instead, we will describe in detail the methodology for creating and testing
those instances.

Although \citet{LoMi10} made available a set of $4200$ generated instances, in
order to have a specific range of values necessary for our objectives we created
our own instances by running their generator.

Specifically, we have created $5$ instances for each combination of values of
the two control parameters of the instance generator. Namely, the number of
lectures ranges in $\{i\cdot 50\, |\, i = 1, \ldots, 24\}$, so that they will be
comprised between $50$ and $1200$, and the percentage of room occupation will be
one of the values $\{50\%, 70\%, 80\%, 90\%\}$. On overall, the full testbed
consists of $480$ instances ($5 \times 24 \times 4$).

After screening them in detail, it appears that not all instances generated are
useful for our parameter tuning purposes. In particular, we have identified the
following instance classes

\begin{description}
    \item[\textbf{Provably infeasible}:] infeasibility is easily proven (e.g., some courses have more lectures than available periods).
    \item[\textbf{Unrealistic room endowment}:] only high cost solutions exist,
      due to the presence of courses with more students than the
      available rooms.
    \item[\textbf{Too hard}:] the solver has never been able to find a feasible solution (given a reasonably long timeout).
    \item[\textbf{Too easy}:] easily solved to cost $0$ by all the solver variants.
\end{description}

Instance belonging to these classes are discarded, since these are features that
generally do not appear in real cases. Furthermore, except for the \emph{too
hard} class, an instance can be easily classified in one of the remaining
classes. In order to have the expected number of instances for any combination
of parameter values, we replace the discarded instances with new ones by
repeating the instance generation procedure.

\subsection{Validation instances}
\label{sec:comp-instances}

The second group comprises those instances that have been used in the
literature. This group consists of the usual set of instances, the so-called
\texttt{comp}, which is composed by $21$ real-world instances that have been
used for ITC-2007.

Over this set we illustrate the solver emerging from the tuning on the first
group, and we compare it with the state-of-the-art methods.

\subsection{Novel instances}
\label{sec:new-instances}

The final set is composed by four families of \emph{novel} instances, which have
been proposed recently so that no (or very few) results are available in the
literature. This set is a candidate to become a new benchmark for future
comparisons.

The first family in this group, contributed by Moritz M\"uhlenthaler, is called
\texttt{Erlangen} and comprises four instances of the course timetabling problem
arising at the University of Erlangen. These instances are considerably larger
than the \texttt{comp} ones and they exhibit a very different structure than
most of the other real-world instances.

The second and third families in this group consist of a set of recent
real-world instances from the University of Udine and of a group of cases from
other Italian universities. These instances have been kindly provided by
EasyStaff S.r.l. (\url{http://www.easystaff.it}), a company specializing in
timetabling solutions. They were collected by the commercial software
EasyAcademy.  We call these two families \texttt{Udine} and \texttt{EasyAcademy}
respectively.

A further set of $7$ instances, called \texttt{DDS}, from other Italian
universities, are available and have has been used occasionally in the
literature.

\subsection{Summary of features}

Table~\ref{tab:features} summarizes the families of instances of the latter two
groups, highlighting some aggregate indicators (i.e., minimum and maximum
values) of the most relevant features.  All instances employed in this work,
with the exception of the training instances, are available from the CB-CTT
website \url{http://satt.diegm.uniud.it/ctt}.

{
\renewcommand{\arraystretch}{1.1}
\renewcommand{\tabcolsep}{4pt}
\begin{table*}
\caption{Minimum and maximum values of the features for the families of instances (\#I: number of instances): courses (C), total lectures (Le), rooms (R), periods (Pe), curricula (Cu), room occupation (RO), average number of conflicts (Co), average teachers availability (Av), room suitability (RS), average daily lectures per curriculum (DL). }
  \centering
\begin{footnotesize}
  \begin{tabular}{lc*{5}{r@{ -- }l}}
  \toprule
    Family & \#I & \multicolumn{2}{c}{C} & \multicolumn{2}{c}{Le} & \multicolumn{2}{c}{R} & \multicolumn{2}{c}{Pe}  & \multicolumn{2}{c}{Cu} \\
    \midrule
\texttt{comp} & 21 & 30 & 131 & 138 & 434 & 5 & 20 & 25 & 36 & 13 & 150 \\
\texttt{DDS} & 7 & 50 & 201 & 146 & 972 & 8 & 31 & 25 & 75 & 9 & 105 \\
\texttt{Udine} & 9 & 62 & 152 & 201 & 400 & 16 & 25 & 25 & 25 & 54 & 101 \\
\texttt{EasyAcademy} & 12 & 50 & 159 & 139 & 688 & 12 & 65 & 25 & 72 & 12 & 65 \\
\texttt{Erlangen}  &  6  &  705 & 850 & 788 & 930 & 110 & 176 & 30 & 30 & 1949 & 3691\\

\midrule
Family & \#I & \multicolumn{2}{c}{RO} & \multicolumn{2}{c}{Co}& \multicolumn{2}{c}{Av}& \multicolumn{2}{c}{RS}& \multicolumn{2}{c}{DL}  \\
\midrule
\texttt{comp} & 21 & 42.6 & 88.9 & 4.7 & 22.1 & 57.0 & 94.2 & 50.2 & 72.4 & 1.5 & 3.9 \\
\texttt{DDS} & 7 & 20.1 & 76.2 & 2.6 & 23.9 & 21.3 & 91.4 & 53.6 & 100.0 & 1.9 & 5.2\\
\texttt{Udine} & 9 & 50.2 & 76.2 & 4.0 & 6.6 & 70.1 & 95.5 & 57.5 & 71.3 & 1.7 & 2.7\\
\texttt{EasyAcademy} & 12 & 17.6 & 52.0 & 4.8 & 22.2 & 55.1 & 100.0 & 41.8 & 70.0 & 2.7 & 7.7\\
\texttt{Erlangen}  &  6  & 15.7 & 25.1 & 3.0 & 3.8 & 66.7 & 71.4 & 45.5 & 56.0 & 0.0 & 0.9\\

\bottomrule
  \end{tabular}
\end{footnotesize}
  \label{tab:features}
\end{table*}
}

\setlength{\tabcolsep}{6pt}

\section{Experimental analysis}
\label{sec:experimental}

We conduct a principled and extensive experimental analysis, aimed at obtaining
a robust tuning of the various parameters of our algorithm, so that its
performances compare favorably with those obtained by other state-of-the-art
methods.

In order to achieve this, we investigate the possible relationships between
instance features (reported in Table~\ref{tab:features}) and ideal
configurations of the solver parameters. The ultimate goal of this study is to
find, for each parameter, either a \emph{fixed value} that works well on a broad
set of instances, or a \emph{procedure} to predict the best value based on
measurable features of each instance. Ideally, the results of this study should
carry over to unseen instances, thus making the approach more general than
classic parameter tuning.  This is, in fact, an attempt to alleviate the effect
of the \emph{No Free Lunch Theorems for Optimization} \citep{WoMa97}, which
state that, for any algorithm (or parameter configuration), any elevated
performance over one class of problems is exactly paid for in performance over
another class.

In the following, we first illustrate the experimental setting and the
statistical methodology employed in this study. We then summarize our findings
and present the experimental results on the \emph{validation} and \emph{novel
instances} compared against the best known results in literature.

\subsection{Design of experiments and experimental setting}
\label{sec:tuning}

Our analysis is based on the $480$ training instances described in Section
\ref{sec:artificial-instances}. The parameter configurations used in the
analysis are sampled from the \emph{Hammersley point set} \citep{HaHW65}, for
the ranges whose bounds are reported in Table~\ref{tab:parameters}. This choice
has been driven by two properties that make this point generation strategy
particularly suitable for parameter tuning.  First, the Hammersley point set is
\emph{scalable}, both with respect to the number of sampled parameter
configurations, and to the dimensions of the sampled space. Second, the sampled
points exhibit \emph{low discrepancy}, i.e., they are space-filling, despite
being random in nature. For these reasons, by sampling the sequence, one can
generate any number of representative combinations of any number of parameters.
Note that the sequence is deterministic, and must be seeded with a list of prime
numbers. Moreover, since the sequence generates points $p \in [0,1]^{n}$, these
values must then be re-scaled in their desired intervals.

All the experiments were generated and executed using \textsc{json2run}
\citep{Urli13} on an Ubuntu Linux 13.04 machine with 16
Intel\textsuperscript{\textregistered} Xeon\textsuperscript{\textregistered} CPU
E5-2660 (2.20 GHz) physical cores, hyper-threaded to 32; each experiment was
dedicated a single virtual core.

\subsection{Exploratory experiments}

In preparation to our analysis, we carried out three preliminary steps.

First, we ran an \emph{F-Race} tuning \citep{BYBS10} of our algorithm over the
training instances with a $95\%$ confidence level, in order to establish a
baseline for further comparisons. The race ended with more than one surviving
configuration, mainly differing for the values of $w_{hard}$ and $cr$, but
giving a clear indication about good values for the other parameters.  This
suggested that setting a specific value for $w_{hard}$ and $cr$, at least within
the investigated intervals, was essentially irrelevant to the performance, and
allowed us to simplify the analysis by fixing $w_{hard} = 100$ and $cr = 0.99$
(see Table~\ref{tab:refined}).  It is worth noticing that, removing an
irrelevant parameter from the analysis has the double benefit of reducing
experimental noise, and allows a finer-grained tuning of the other parameters,
at the same computational cost.

Secondly, we tested all the sampled parameter configurations against the whole set
of training instances. This allowed us to further refine our study in two ways.
First, we realized that our initial estimates over the parameter intervals were
too conservative, encompassing low-performance areas of the parameters space. A
notable finding was that, on the whole set of training instances, a golden spot
for $sr$ was around $0.43$. We thus fixed this parameter as well, along with
$w_{hard}$ and $cr$. Table~\ref{tab:refined} summarizes the whole parameter
space after this preliminary phase (parameters in boldface have not been fixed
in this phase, and are the subject of the following analysis). Second, by
running a Kruskal-Wallis test \citep[see][]{HoWD2013} with significance level
$10\%$ on the dependence of cost distribution on parameter values, we realized
that some of the instances were irrelevant to our analysis, and we therefore
decided to drop them, and to limit our study to the $314$ significant ones.

Finally, we sampled $20$ parameter configurations from the remaining
3-dimensional ($T_{0}$, $\rho$, $T_{min}$) Hammersley point space (see
Table~\ref{tab:percent}), repeated the race with fewer parameters, and found a
single winning configuration, corresponding to configuration \#11 in
Table~\ref{tab:percent}. In addition, we performed $10$ independent runs of each
parameter configuration on every instance. This is the data upon which the rest
of our experimental analysis is based.

{
    \renewcommand{\arraystretch}{1.1}
    \renewcommand{\tabcolsep}{2pt}

    \begin{table}
        \caption{Revised intervals for investigated parameters\label{tab:refined}.}
        \centering
        \begin{tabular}{llc}
        \toprule
            Parameter                 & Symbol & Range     \\
            \midrule
            \textbf{Starting temperature}               & \textbf{$T_{0}$}        & \textbf{$[1, 40]$}         \\
            \textbf{Neighbors accepted ratio ($n_a/n_s$)} & \textbf{$\rho$}        & \textbf{$[0.034, 0.05]$}         \\
            Cooling rate                       & $cr$           & $\{0.99\}$       \\
            Hard constraints weight            & $w_{hard}$     & $\{100\}$            \\
            Neighborhood swap rate             & $sr$           & $\{0.43\}$         \\
            \textbf{Expected minimum temperature}       & \textbf{$T_{min}$}      & \textbf{$[0.015,0.21]$}   \\
    	\bottomrule
        \end{tabular}
    \end{table}
}

\subsection{Statistical methodology}
\label{sec:statistical}

In order to build a model to \emph{predict} the parameter values for each
instance, we proceed in two stages. In the first stage, for each of the selected
training instances we identify the effective parameter configurations.  The
second stage consists in learning a rule to associate the parameter
configurations to the features. In principle, each of the two stages can be
performed using various strategies. Some compelling choices are described in the
following.

\subsubsection{Identification of the effective parameter configurations}
\label{sec:statistical:1}

For each instance, we aim at identifying which of the $20$ experimental
configurations perform well.  We start by selecting the \emph{best-performing
configuration}, defined here as the configuration with the smallest median cost
(as described in Section \ref{sec:cost}). Other robust estimates of location
could be employed for the task, such as the Hodges-Lehmann estimator
\citep[e.g.,][\S 3.2]{HoWD2013}, but the median is the simplest possibility.
Then, a series of Wilcoxon rank-sum tests are run to identify the configurations
which can be taken as equivalent to the best-performing one. For the latter
task, it is crucial to adjust for multiple testing, as a plurality of
statistical tests are carried out. Here we control the \emph{false discovery rate}
(FDR) \citep{Benjamini95}, thus adopting a more powerful approach
than classical alternatives which control the family-wise error rate, such as
Bonferroni or Holm methods \citep[see][Chapter 18]{HaTF2009}.  In short, we
accept to wrongly declare some of the configurations as less performing than
the best one, rather than strictly controlling the proportion of such mistakes,
ending up declaring nearly all the configurations as equivalent to the
best-performing one for several instances. In this study we set a FDR threshold
equal to $0.10$.

The result of this first stage is a $314 \times 20$ binary matrix; the $i$-th
row of such a matrix contains the value $1$ for those configurations equivalent to
the best-performing one on the $i$-th instance, and $0$ otherwise.  Table
\ref{tab:percent} summarizes the column averages for such a matrix, thus
reporting the percentages of ``good performances'' for each of the experimental
configurations. The striking result is that some of the configurations, such as
\#3, 4, 5, 10 and 11, perform well in the vast majority of the instances.  We
will return on this point in Section \ref{sec:comparison-frace}.

{
\renewcommand{\arraystretch}{1.1}
    \renewcommand{\tabcolsep}{2pt}
    \begin{table}
        \caption{Percentage of ``good performances'' for the 20 experimental configurations \label{tab:percent} (i.e., the percentage of instances in which the given configuration is equivalent to the best one).}
        \centering
        \begin{tabular}{ccccc}
        \toprule
     Configuration & \textbf{$T_{0}$} & \textbf{$T_{min} \times 100$} & \textbf{$\rho \times 100$} & \% Good performance\\
      \hline
    1 & 21.72 & 20.56 & 4.76 & 54.8 \\
      2 & 2.22 & 18.57 & 4.68 & 20.7 \\
      3 & 20.50 & 17.00 & 3.48 & 93.0 \\
      4 & 25.38 & 19.67 & 3.80 & 87.6 \\
      5 & 22.94 & 15.22 & 4.12 & 89.5 \\
      6 & 27.81 & 17.89 & 4.44 & 79.3 \\
      7 & 3.44 & 20.35 & 4.04 & 24.5 \\
      8 & 32.69 & 19.22 & 4.28 & 82.5 \\
      9 & 31.47 & 17.45 & 4.92 & 75.5 \\
      10 & 35.13 & 18.34 & 3.96 & 93.3 \\
      11 & 30.25 & 15.67 & 3.64 & 93.9 \\
      12 & 37.56 & 16.55 & 4.60 & 83.4 \\
      13 & 5.88 & 17.68 & 3.72 & 50.3 \\
      14 & 7.09 & 19.43 & 5.00 & 32.2 \\
      15 & 8.31 & 15.88 & 4.36 & 58.3 \\
      16 & 10.75 & 19.00 & 3.56 & 69.7 \\
      17 & 13.19 & 17.23 & 4.20 & 66.2 \\
      18 & 11.97 & 15.45 & 4.84 & 57.0 \\
      19 & 18.06 & 19.89 & 4.52 & 58.6 \\
      20 & 15.63 & 16.34 & 3.88 & 85.0 \\
     \bottomrule
        \end{tabular}
    \end{table}
}

\subsubsection{Prediction of optimal configurations based on
  measurable features}

Once the performances of each parameter configuration have been identified for
each instance, we include instance features in the process. We tackle the
problem as a classification problem, and use the instance features to predict
the outcome of each of the $20$ binary variables reporting the performance of
the different experimental configuration.  In other words, we build $20$ different
classifiers, each using the instance features to return the probability of good
performance of the experimental configuration. The configuration achieving the
highest probability is then identified as the outcome of the classification
process. The idea is that by making use of the instance features we have a
chance to improve over the majority class rule, that simply picks up the
configuration with the highest success rate in Table \ref{tab:percent}, namely
configuration \#11.  In the worst case, where no feature is able to provide some
useful information for performing the classification, we can revert to
the majority class rule, which can be considered as a baseline strategy.  In our
experimental setting, where the best-performing configuration has a success rate
as high as $93.9\%$, getting a significant improvement is, therefore, not
straightforward.

Among the possible classification methods, we choose the random forests method
\citep{Breiman01}. This classification method is an extension of
classification trees, that perform rather well compared to alternative methods,
and for which efficient software is available.  We use the implementation
provided by the R package \texttt{randomForest} \citep{LiWi02}, using $5'000$
trees for growing the forest.

We train the different classifiers by including all the available features
reported in Table \ref{tab:features}, with the exception of periods (Pe), which
was assumed to be constant across the $314$ training instances selected, courses
(C) and rooms (R), which were essentially duplicating the information content of
total lectures (Le). A total number of seven features were therefore used in the
classification process.

We estimate the additional gain provided by feature-based classification by
means of $10$-fold cross validation, which provides an estimate of the
classification error \citep[][\S 7.10]{HaTF2009}. We obtain a classification
accuracy for the method based on random forests equal to $95.5\%$, which is better
than simpler methods such as logistic regression or generalized additive models,
providing a classification accuracy equal to $94.3\%$ an $93.9\%$ respectively.
Although we use random forests as a black box classifier, one of the good
properties of such methodology is that it also returns some measures of feature
importance. Based on the prediction strength \citep[][\S 15.3.2]{HaTF2009}, we
found that for $10$ out of the $20$ classifiers, total lectures (Le) is the most
important variable, and in the remaining $10$ classifiers that role is played by
the average number of conflicts (Co). These two features can then be deemed as the
most important for selecting algorithm parameters.

\subsection{Comparison with other approaches}

In order to validate the quality of our approach, we compared its results
against the best ones in literature using the ITC-2007 rules and instances.  For
the sake of fairness, we do not include in the comparison results that are
obtained by allotting a higher runtime, for example those of \cite{AsNi14}, who
use $10'000$ to $100'000$ seconds instead of about $300-500$ seconds as
established by the competition rules.

Table~\ref{tab:results-comp-avg} shows the average of $31$ repetitions of the
algorithm, in which we have highlighted in boldface the lowest average costs. We
report the results obtained with both feature-based tuning (FBT) and standard
F-Race tuning (further discussed in Section~\ref{sec:comparison-frace}). The
figures show that our approach matches or outperforms the state-of-the-art
algorithms in more than half of the instances, also improving on our previous
results \citep{BeDS12}.

This outcome is particularly significant because unlike previous approaches we
have not involved the validation instances in the tuning process, thus avoiding
the classical overtuning phenomenon.

\subsection{Comparison with the F-Race baseline}
\label{sec:comparison-frace}

The figures in Table~\ref{tab:results-comp-avg} reveal that the feature-based
tuning outperforms the F-Race approach on most instances in terms of average
results. A more precise assessment reveals that the two methods are broadly
comparable, as the Wilcoxon rank-sum test is significant at the $5\%$ level (in
favor of FBT) only on one instance (\texttt{comp14}): consequently, any sensible
global test would not flag any significant difference between the two methods.
At any rate, we can safely say that the FBT method is no worse than F-Race, and
occasionally can be better.

It is worth noting that the above conclusion is something to be expected.
Indeed, the parameter configuration selected by F-Race is \#11, which
corresponds to the best-performing configuration over the training instances
(see Table \ref{tab:percent}).  Strictly speaking, the equivalence between
F-Race and the proposed FBT when the features are not informative is not
perfect, as F-Race uses a different adjustment for multiple testing compared to
the FDR method adopted for FBT, but nevertheless it is hardly surprising that
the configuration surviving the race is the one with the highest percentage of
good performances.

\begin{figure}[htbp]
\begin{center}
 \includegraphics[width=.7\textwidth]{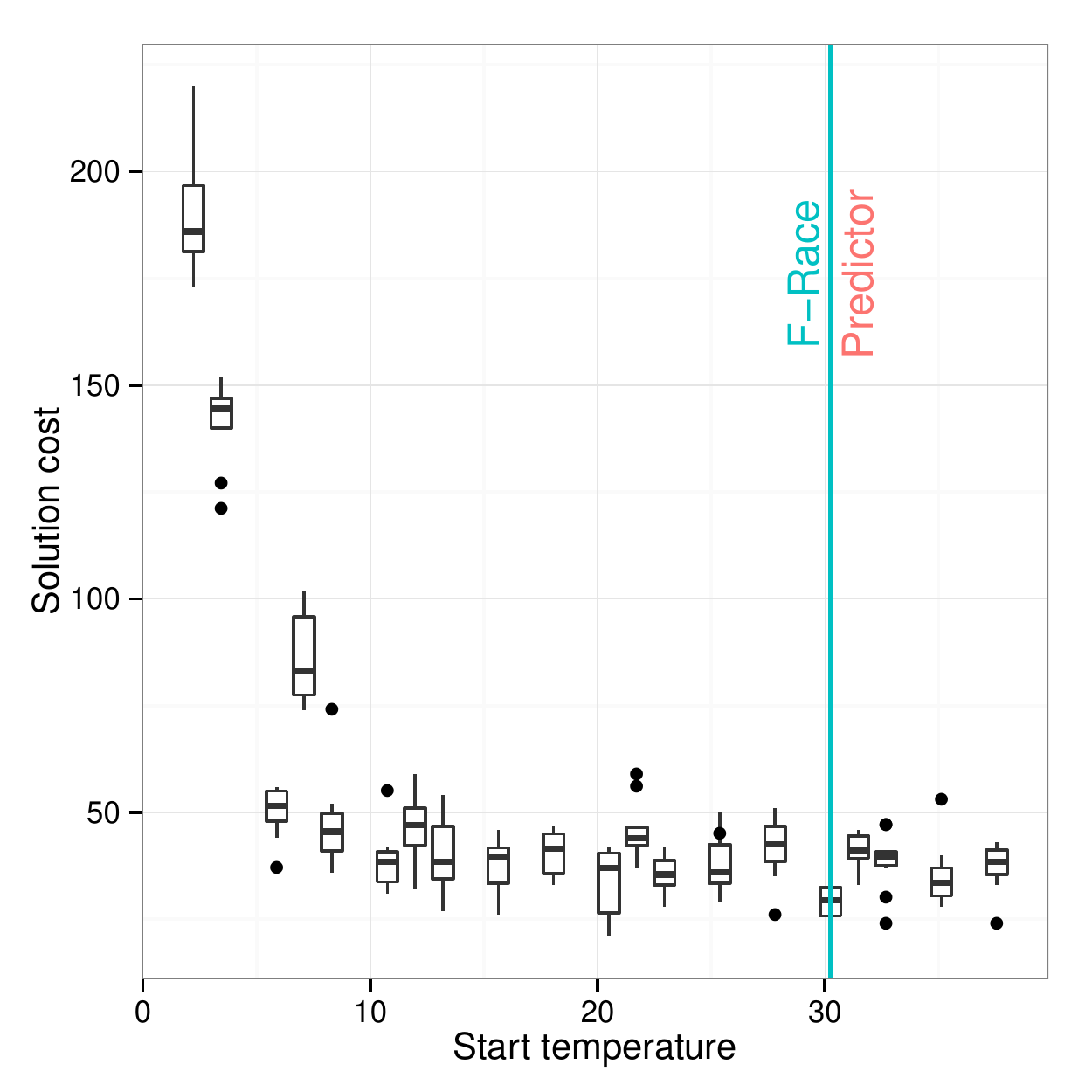}
\caption{Relation between $T_{0}$ and cost distribution on a single
     training instance with $Le=1200$. We show the ideal $T_{0}$ found
     by F-Race and by our feature-based predictor.}
\label{fig:correlation}
\end{center}
\end{figure}

At this point, it makes sense to look for an explanation of why some
configurations could achieve quite high success rates in the training phase. By
looking at Figure \ref{fig:correlation}, which is representative of a large
portion of the training instances, it is possible to glean some information. The
parameter space (in this case for $T_{0}$) is split in two well-separated parts.
One part (the leftmost in Figure \ref{fig:correlation}) yields poor results,
while any choice of values inside the other part is reasonably safe. In our
scenario, the portion of the parameter space leading to poor results is narrow,
while the portion leading to better results is broader. This suggests that the
chosen algorithm is robust with respect to parameter choice, at least for this
problem domain. As a consequence, it is possible to find some parameter
configurations that work consistently well across a large set of instances.

{

    \begin{table*}[htdp]
    \caption{Best results and comparison with other approaches over the \emph{validation} instances. Values are averages over multiple runs of the algorithms.}
    \begin{scriptsize}
    \begin{center}

    \begin{tabular}{l|*{4}{r}|r|r}
\toprule
                    & \citeauthor{Mull09}   &  \citeauthor{LuHa09}     &  \citeauthor{ATMM10}  &  \citeauthor{BeDS12}    & us (FBT)       & us (F-Race)  \\  
\midrule
\texttt{comp01}  & \textbf{5.00} & \textbf{5.00} & \textbf{5.00} & \textbf{5.00} & 5.23 & 5.16 \\
\texttt{comp02}  & 61.30 & 60.60 & 53.90 & \textbf{51.60} & 52.94 & 53.42 \\
\texttt{comp03}  & 94.80 & 86.60 & 84.20 & 82.70 & \textbf{79.16} & 80.48 \\
\texttt{comp04}  & 42.80 & 47.90 & 51.90 & 47.90 & 39.39 & \textbf{39.29} \\
\texttt{comp05}  & 343.50 & \textbf{328.50} & 339.50 & 333.40 & 335.13 & 329.06 \\
\texttt{comp06}  & 56.80 & 69.90 & 64.40 & 55.90 & \textbf{51.77} & 53.35 \\
\texttt{comp07}  & 33.90 & 28.20 & \textbf{20.20} & 31.50 & 26.39 & 28.45 \\
\texttt{comp08}  & 46.50 & 51.40 & 47.90 & 44.90 & 43.32 & \textbf{43.06} \\
\texttt{comp09}  & 113.10 & 113.20 & 113.90 & 108.30 & \textbf{106.10} & \textbf{106.10} \\
\texttt{comp10}  & \textbf{21.30} & 38.00 & 24.10 & 23.80 & 21.39 & 21.71 \\
\texttt{comp11}  & \textbf{0.00} & \textbf{0.00} & \textbf{0.00} & \textbf{0.00} & \textbf{0.00} & \textbf{0.00} \\
\texttt{comp12}  & 351.60 & 365.00 & 355.90 & 346.90 & \textbf{336.84} & 338.39 \\
\texttt{comp13}  & 73.90 & 76.20 & \textbf{72.40} & 73.60 & 73.39 & 73.65 \\
\texttt{comp14}  & 61.80 & 62.90 & 63.30 & 60.70 & \textbf{58.16} & 59.71 \\
\texttt{comp15}  & 94.80 & 87.80 & 88.00 & 89.40 & \textbf{78.19} & 79.10 \\
\texttt{comp16}  & 41.20 & 53.70 & 51.70 & 43.00 & \textbf{38.06} & 39.19 \\
\texttt{comp17}  & 86.60 & 100.50 & 86.20 & 83.10 & \textbf{77.61} & 78.84 \\
\texttt{comp18}  & 91.70 & 82.60 & 85.80 & 84.30 & \textbf{81.10} & 83.29 \\
\texttt{comp19}  & 68.80 & 75.00 & 78.10 & 71.20 & \textbf{66.77} & 67.13 \\
\texttt{comp20}  & \textbf{34.30} & 58.20 & 42.90 & 50.60 & 46.13 & 45.94 \\
\texttt{comp21}  & 108.00 & 125.30 & 121.50 & 106.90 & 103.32 & \textbf{102.19} \\
\bottomrule
    \end{tabular}
    \end{center}
    \end{scriptsize}
    \label{tab:results-comp-avg}
    \end{table*}%
}

\subsection{Results on the novel instances}

In Tables~\ref{tab:results-test-avg}--\ref{tab:results-erlangen} we show the
results obtained by $31$ repetitions of our algorithm on the instances of the
\emph{novel} group of families, for future comparison with these instances.  The
results are reported both for the F-Race and the feature-based tuning.

Across all the $28$ instances of Table~\ref{tab:results-test-avg}, only in two
cases there was a significant difference at the $5\%$ level (in favor of FBT,
namely for the \texttt{Udine2} and \texttt{Udine3}), so that the two methods are
largely equivalent in performance.

Table~\ref{tab:results-test-avg} does not include the instances of the
\texttt{Erlangen} family because this family shows a very particular structure,
caused by the peculiar way in which the original timetabling problem is
represented.  In particular, the average number of lectures (Le) is very close
to the average number of courses (C) (see Table~\ref{tab:features}), suggesting
that, in almost all cases, courses are composed of a single lecture.  Similarly,
the number of curricula (Cu), which is typically smaller than the number of
courses, is instead one order of magnitude larger, possibly because each
curriculum represents the choice of a single student. This kind of mapping is
closer in structure to the one used in the \pectt{} formulation, which has been
addressed in \cite{CDGS12}.

Therefore, for the \texttt{Erlangen} instances we decided to present also the
results of a new F-Race tuning done \emph{ad hoc} for these cases, using
different ranges for the parameter values. In fact, it turned out that the
winning parameter configuration is $T_{0} = 100$, $T_{min} = 0.36$, $\rho =
0.23$, $sr = 0.43$ and $w_{hard} = 500$, which is considerably different from
the ones used in the previous analyses.

For comparison, we report in Table~\ref{tab:results-erlangen} also the results
obtained by re-running the solver by \cite{Mull09}, which is available online.
Given that these instances are larger, we used a longer timeout of $600$ seconds.

It is worth mentioning that, although the different semantics of the involved
features for this family of instances, our feature-based predictor was the only
one able to obtain always feasible solutions, proving its robustness and wide
range applicability.  Indeed, M\"uller's solver, and the F-Race and Ad Hoc
F-Race-tuned versions of our solver, found feasible solutions for $81\%$, $21\%$
and $74\%$ of the runs, respectively.

Looking at the average values reported in Table~\ref{tab:results-erlangen}, we
notice that whenever a feasible solution was found, our F-Race tuned solver
outperforms all the other methods on all $6$ instances, including
\citeauthor{Mull09}'s.

{
    \renewcommand{\arraystretch}{1.1}
       \renewcommand{\tabcolsep}{14pt}

    \begin{table}[htdp]
    \caption{Our results on the \emph{novel} instances.}
    \begin{scriptsize}
    \begin{center}
    \begin{tabular}{l|rr|rr}

\toprule
                        & \multicolumn{2}{c|}{us (FBT)}& \multicolumn{2}{c}{us (F-Race)}       \\
                                & \emph{best} & \emph{avg}           & \emph{best} & \emph{avg}                           \\
\midrule
\texttt{Udine1} & 5 & 13.29 & 7 & 12.45 \\
\texttt{Udine2} & 14 & 19.26 & 11 & 21.42 \\
\texttt{Udine3} & 3 & 8.52 & 5 & 10.23 \\
\texttt{Udine4} & 64 & 66.16 & 64 & 66.32 \\
\texttt{Udine5} & 0 & 3.13 & 0 & 2.97 \\
\texttt{Udine6} & 0 & 0.35 & 0 & 0.19 \\
\texttt{Udine7} & 0 & 2.03 & 0 & 2.32 \\
\texttt{Udine8} & 34 & 39.26 & 33 & 39.10 \\
\texttt{Udine9} & 23 & 29.84 & 26 & 29.90 \\
\midrule
\texttt{EasyAcademy01} & 65 & 65.26 & 65 & 65.03 \\
\texttt{EasyAcademy02} & 0 & 0.06 & 0 & 0.03 \\
\texttt{EasyAcademy03} & 2 & 2.06 & 2 & 2.10 \\
\texttt{EasyAcademy04} & 0 & 0.35 & 0 & 0.10 \\
\texttt{EasyAcademy05} & 0 & 0.00 & 0 & 0.00 \\
\texttt{EasyAcademy06} & 5 & 5.13 & 5 & 5.06 \\
\texttt{EasyAcademy07} & 0 & 0.48 & 0 & 0.32 \\
\texttt{EasyAcademy08} & 0 & 0.00 & 0 & 0.00 \\
\texttt{EasyAcademy09} & 4 & 5.16 & 4 & 4.74 \\
\texttt{EasyAcademy10} & 0 & 0.03 & 0 & 0.06 \\
\texttt{EasyAcademy11} & 0 & 2.90 & 0 & 3.23 \\
\texttt{EasyAcademy12} & 4 & 4.03 & 4 & 4.00 \\
\midrule
\texttt{DDS1} & 85 & 110.26 & 91 & 106.55 \\
\texttt{DDS2} & 0 & 0.00 & 0 & 0.00 \\
\texttt{DDS3} & 0 & 0.00 & 0 & 0.00 \\
\texttt{DDS4} & 18 & 21.13 & 17 & 20.55 \\
\texttt{DDS5} & 0 & 0.00 & 0 & 0.00 \\
\texttt{DDS6} & 5 & 9.87 & 4 & 10.35 \\
\texttt{DDS7} & 0 & 0.00 & 0 & 0.00 \\
\bottomrule

    \end{tabular}
    \end{center}
    \end{scriptsize}
    \label{tab:results-test-avg}
    \end{table}%
    }

{
    \renewcommand{\arraystretch}{1.1}
       \renewcommand{\tabcolsep}{4pt}

    \begin{table}[htdp]
    \caption{Our results on the \emph{Erlangen} instances.}
    \begin{scriptsize}
    \begin{center}
    \begin{tabular}{l|rrc|rrc|rrc|rrc}

    \toprule
                        & \multicolumn{3}{c|}{\citet{Mull09}}   & \multicolumn{3}{c|}{us (FBT)}   & \multicolumn{3}{c}{us (F-Race)}   & \multicolumn{3}{c}{us (\emph{ad hoc} F-Race)}  \\
                                & \emph{best} & \emph{avg}      & \emph{feas}   & \emph{best} & \emph{avg}      & \emph{feas}    & \emph{best} & \emph{avg}          & \emph{feas}     & \emph{best} & \emph{avg}          & \emph{feas}             \\
    \midrule

Erlangen-2011-2 & 5569 & 5911.10 & 32\% & 5444 & 6568.42 & 100\% & -- & -- & 0\% & 4680 & 4971.80 & 56\% \\
Erlangen-2012-1 & 8059 & 8713.48 & 100\% & 7991 & 9130.87 & 100\% & 8174 & 8604.77 & 71\% & 7519 & 7856.85 & 100\% \\
Erlangen-2012-2 & 11267 & 13367.53 & 61\% & 13693 & 16183.74 & 100\% & -- & -- & 0\% & 9587 & 10130.33 & 7\% \\
Erlangen-2013-1 & 7993 & 8878.19 & 94\% & 7445 & 9299.94 & 100\% & 7745 & 8174.33 & 29\% & 7159 & 7825.55 & 95\% \\
Erlangen-2013-2 & 8809 & 19953.81 & 100\% & 8846 & 11273.97 & 100\% & -- & -- & 0\% & 8329 & 8844.34 & 86\% \\
Erlangen-2014-1 & 6728 & 7978.42 & 100\% & 6601 & 8251.13 & 100\% & 6418 & 7080.75 & 26\% & 6144 & 6525.05 & 100\% \\

\bottomrule
   \end{tabular}
    \end{center}
    \end{scriptsize}
    \label{tab:results-erlangen}
    \end{table}%
}

\subsection{Results on the \texttt{comp} instances for the other problem formulations}
\label{sec:other-formulations}

Finally, in order to given a more comprehensive picture of the solver
robustness, we report its results on other formulations of the problem with
different constraints and objectives. In particular, we consider the
formulations \texttt{UD3}, \texttt{UD4}, and \texttt{UD5}, proposed by
\citet{BDDS12}, that focus on student load, double lectures presence, and travel
costs, respectively.

\begin{table}[htdp]
   \begin{scriptsize}
         \caption{Our results on the \texttt{comp} instances for the UD3 formulation.}
      \label{tab:results-UD3}
    \begin{center}
      \begin{tabular}{l|rr|rr|r}
        \toprule
        & \multicolumn{2}{c|}{us (FBT)}   & \multicolumn{2}{c|}{us (F-Race)}   & \multicolumn{1}{c}{\citet{MSTIS13}}    \\
        Instance &  \emph{best} & \emph{avg} & \emph{best} & \emph{avg} & z \\
        \midrule
\texttt{comp01} & 8 & 8.00 & 8 & 8.00 & 10 \\
\texttt{comp02} & 15 & 22.13 & 14 & 24.06 & 12 \\
\texttt{comp03} & 29 & 35.71 & 27 & 34.00 & 147 \\
\texttt{comp04} & 2 & 2.06 & 2 & 2.19 & 2 \\
\texttt{comp05} & 271 & 331.55 & 271 & 329.29 & 1232 \\
\texttt{comp06} & 8 & 11.42 & 8 & 11.94 & 8 \\
\texttt{comp07} & 0 & 2.00 & 0 & 1.81 & 0 \\
\texttt{comp08} & 2 & 3.61 & 2 & 4.19 & 2 \\
\texttt{comp09} & 8 & 9.87 & 8 & 11.19 & 8 \\
\texttt{comp10} & 0 & 1.74 & 0 & 1.35 & 0 \\
\texttt{comp11} & 0 & 0.00 & 0 & 0.00 & 0 \\
\texttt{comp12} & 54 & 76.23 & 57 & 77.94 & 1281 \\
\texttt{comp13} & 22 & 24.77 & 22 & 25.29 & 63 \\
\texttt{comp14} & 0 & 0.13 & 0 & 0.19 & 0 \\
\texttt{comp15} & 18 & 24.06 & 16 & 24.19 & 118 \\
\texttt{comp16} & 4 & 5.81 & 4 & 5.87 & 4 \\
\texttt{comp17} & 12 & 14.35 & 12 & 14.39 & 12 \\
\texttt{comp18} & 0 & 0.84 & 0 & 0.39 & 0 \\
\texttt{comp19} & 24 & 30.65 & 26 & 33.13 & 93 \\
\texttt{comp20} & 0 & 5.42 & 0 & 4.23 & 0 \\
\texttt{comp21} & 12 & 17.87 & 10 & 17.29 & 6 \\
\bottomrule
   \end{tabular}
    \end{center}
    \end{scriptsize}
    \end{table}%

      \begin{table}[htdp]
    \begin{scriptsize}
    \caption{Our results on the  \texttt{comp} instances for the UD4 formulation.}
     \label{tab:results-UD4}
    \begin{center}
    \begin{tabular}{l|rr|rr|r}
  \toprule
                 & \multicolumn{2}{c|}{us (FBT)}   & \multicolumn{2}{c|}{us (F-Race-F2)}   & \multicolumn{1}{c}{\citet{MSTIS13}}    \\
 Instance &  \emph{best} & \emph{avg} & \emph{best} & \emph{avg} & z \\
  \midrule
\texttt{comp01} & 6 & 6.00 & 6 & 6.00 & 9 \\
\texttt{comp02} & 26 & 31.19 & 28 & 32.81 & 107 \\
\texttt{comp03} & 501 & 547.55 & 512 & 551.77 & 474 \\
\texttt{comp04} & 13 & 14.48 & 13 & 14.55 & 13 \\
\texttt{comp05} & 247 & 258.71 & 248 & 260.81 & 584 \\
\texttt{comp06} & 14 & 18.13 & 15 & 19.00 & 39 \\
\texttt{comp07} & 6 & 8.45 & 5 & 8.06 & 3 \\
\texttt{comp08} & 15 & 16.84 & 15 & 16.74 & 15 \\
\texttt{comp09} & 38 & 40.35 & 38 & 40.48 & 122 \\
\texttt{comp10} & 6 & 9.74 & 6 & 10.39 & 3 \\
\texttt{comp11} & 0 & 0.00 & 0 & 0.00 & 0 \\
\texttt{comp12} & 98 & 108.65 & 100 & 111.35 & 479 \\
\texttt{comp13} & 41 & 43.65 & 41 & 44.16 & 109 \\
\texttt{comp14} & 16 & 17.65 & 16 & 18.03 & 18 \\
\texttt{comp15} & 30 & 34.81 & 29 & 35.13 & 129 \\
\texttt{comp16} & 12 & 14.35 & 11 & 13.97 & 7 \\
\texttt{comp17} & 25 & 28.13 & 25 & 28.29 & 58 \\
\texttt{comp18} & 25 & 27.84 & 24 & 27.52 & 93 \\
\texttt{comp19} & 33 & 37.52 & 33 & 37.23 & 123 \\
\texttt{comp20} & 11 & 14.29 & 10 & 14.03 & 168 \\
\texttt{comp21} & 36 & 41.23 & 36 & 40.81 & 121 \\
\bottomrule
   \end{tabular}
    \end{center}
    \end{scriptsize}
    \end{table}%

    \begin{table}[htdp]
    \begin{scriptsize}
    \caption{Our results on the  \texttt{comp} instances for the UD5 formulation.}
     \label{tab:results-UD5}
    \begin{center}
    \begin{tabular}{l|rr|rr|r}
  \toprule
                 & \multicolumn{2}{c|}{us (FBT)}   & \multicolumn{2}{c|}{us (F-Race-F2)}   & \multicolumn{1}{c}{\citet{MSTIS13}}    \\
 Instance &  \emph{best} & \emph{avg} & \emph{best} & \emph{avg} & z \\
  \midrule
\texttt{comp01} & 11 & 11.06 & 11 & 11.13 & 45 \\
\texttt{comp02} & 135 & 160.19 & 139 & 161.45 & 714 \\
\texttt{comp03} & 140 & 165.58 & 140 & 167.65 & 523 \\
\texttt{comp04} & 56 & 67.42 & 57 & 67.74 & 215 \\
\texttt{comp05} & 568 & 619.29 & 587 & 633.58 & 2753 \\
\texttt{comp06} & 82 & 102.45 & 88 & 102.39 & 747 \\
\texttt{comp07} & 43 & 55.77 & 37 & 53.10 & 910 \\
\texttt{comp08} & 64 & 72.87 & 64 & 72.68 & 212 \\
\texttt{comp09} & 154 & 166.97 & 152 & 165.90 & 428 \\
\texttt{comp10} & 68 & 86.42 & 69 & 88.29 & 633 \\
\texttt{comp11} & 0 & 0.00 & 0 & 0.00 & 0 \\
\texttt{comp12} & 471 & 515.32 & 471 & 522.03 & 2180 \\
\texttt{comp13} & 148 & 160.39 & 147 & 159.32 & 488 \\
\texttt{comp14} & 97 & 107.06 & 96 & 109.39 & 541 \\
\texttt{comp15} & 173 & 189.84 & 170 & 191.65 & 656 \\
\texttt{comp16} & 95 & 109.26 & 93 & 109.58 & 914 \\
\texttt{comp17} & 154 & 165.19 & 152 & 169.00 & 818 \\
\texttt{comp18} & 133 & 143.19 & 136 & 142.65 & 509 \\
\texttt{comp19} & 123 & 140.52 & 121 & 137.19 & 619 \\
\texttt{comp20} & 124 & 146.10 & 123 & 147.61 & 2045 \\
\texttt{comp21} & 151 & 168.00 & 145 & 170.52 & 651 \\
 \bottomrule
   \end{tabular}
    \end{center}
    \end{scriptsize}
    \end{table}%

Tables~\ref{tab:results-UD3}--\ref{tab:results-UD5} present average and best
results for the configurations obtained by both tuning methods, on the three
formulations.  Those tables also show the results by \citet{MSTIS13}, which are,
to our knowledge, the only published results on these formulations.

The comparison reveals that again FBT and F-Race perform similarly, and
that in general, they perform better than the solver by \citet{MSTIS13}.

\section{Conclusions and future work}

In this paper we have proposed a simple yet effective SA approach to the
\cbctt{} problem.
Moreover, we performed a comprehensive statistical analysis to identify the
relationship between instance features and search method parameters. The
outcome of this analysis makes it possible to set the parameters for unseen
instances on the basis of a simple inspection of the instance itself.

The results of the feature-based tuned algorithm on a testbed of
\emph{validation} instances allows us to improve our previous results
\citep{BeDS12} and outperform the results in the literature on $10$ instances
out of $21$.

The results of this work support the conclusion that instance features may
provide useful information for tuning the parameters of the solver. Therefore, a
sensible direction for future investigation may focus on refining this approach
by exploring novel features obtained as functions, possibly complex in nature,
of the original features.

In the future, we plan to investigate new versions of SA that could be applied
to this problem. For example, we plan to consider the version that includes
reheating.  In addition, we plan to investigate other metaheuristics techniques
for the problem through an analogous statistical analysis.  Finally, we aim at
applying this search and tuning technique to other timetabling problems.

\end{document}